\documentclass[runningheads]{llncs}
\usepackage[T1]{fontenc}
\usepackage{graphicx}
\usepackage{subfig}
\usepackage{hyperref}
\begin{document}
\title{Feedback-driven object detection and iterative model improvement for accurate annotations}
\titlerunning{Feedback-driven object detection and iterative model improvement}
%
%
\author{Sönke Tenckhoff$^{*}$\orcidID{0009-0002-5341-4314} \and
Mario Koddenbrock$^{*}$\orcidID{0000-0003-3327-7404} \and
Erik Rodner\orcidID{0000-0002-3711-1498}}

\authorrunning{Tenckhoff, Koddenbrock, Rodner}
%
\institute{KI-Werkstatt/Fachbereich 2, University of Applied Sciences Berlin,\\ Wilhelminenhofstr. 75A, 12459 Berlin, Germany\\
\email{firstname.lastname@htw-berlin.de}
}

\maketitle   

\begin{center}
\footnotesize{*These authors contributed equally}
\end{center}

\begin{abstract}

Automated object detection has become increasingly valuable across diverse applications, yet efficient, high-quality annotation remains a persistent challenge. In this paper, we present the development and evaluation of a platform designed to interactively improve object detection models. The platform allows uploading and annotating images as well as fine-tuning object detection models. Users can then manually review and refine annotations, further creating improved snapshots that are used for automatic object detection on subsequent image uploads---a process we refer to as \textit{semi-automatic annotation} resulting in a significant gain in annotation efficiency.

Whereas iterative refinement of model results to speed up annotation has become common practice, we are the first to quantitatively evaluate its benefits with respect to time, effort, and interaction savings. Our experimental results show clear evidence for a significant time reduction of up to \textbf{53\%} for semi-automatic compared to manual annotation. Importantly, these efficiency gains did not compromise annotation quality, while matching or occasionally even exceeding the accuracy of manual annotations. These findings demonstrate the potential of our lightweight annotation platform for creating high-quality object detection datasets and provide best practices to guide future development of annotation platforms. 


The platform is open-source, with the frontend and backend repositories available on GitHub\footnote{\url{https://github.com/ml-lab-htw/iterative-annotate}}. To support the understanding of our labeling process, we have created an explanatory video demonstrating the methodology using microscopy images of E. coli bacteria as an example. The video is available on YouTube\footnote{\url{https://www.youtube.com/watch?v=CM9uhE8NN5E}}.

\end{abstract}

\section{Introduction}

Object detection has become a critical component in various computer vision applications, including but not limited to autonomous driving \cite{takumi2017multispectral}, surveillance\cite{fedorov2019_traffic_flow}, robotics \cite{robotmanipulation}, microscopy \cite{midtvedt2022single}, and manufacturing \cite{ahmad2022deep}. These applications rely on high-quality labeled datasets to train and fine-tune models. Traditionally, creating such datasets requires manually tagging images—a process that is both labor-intensive and prone to human error. As object detection models continue to evolve, so do the strategies for efficiently generating these annotated datasets, which has become a key research focus \cite{wu2023label,GARCIAAGUILAR202345}.

\begin{figure}[t!]
    \centering
    \includegraphics[width=0.95\linewidth]{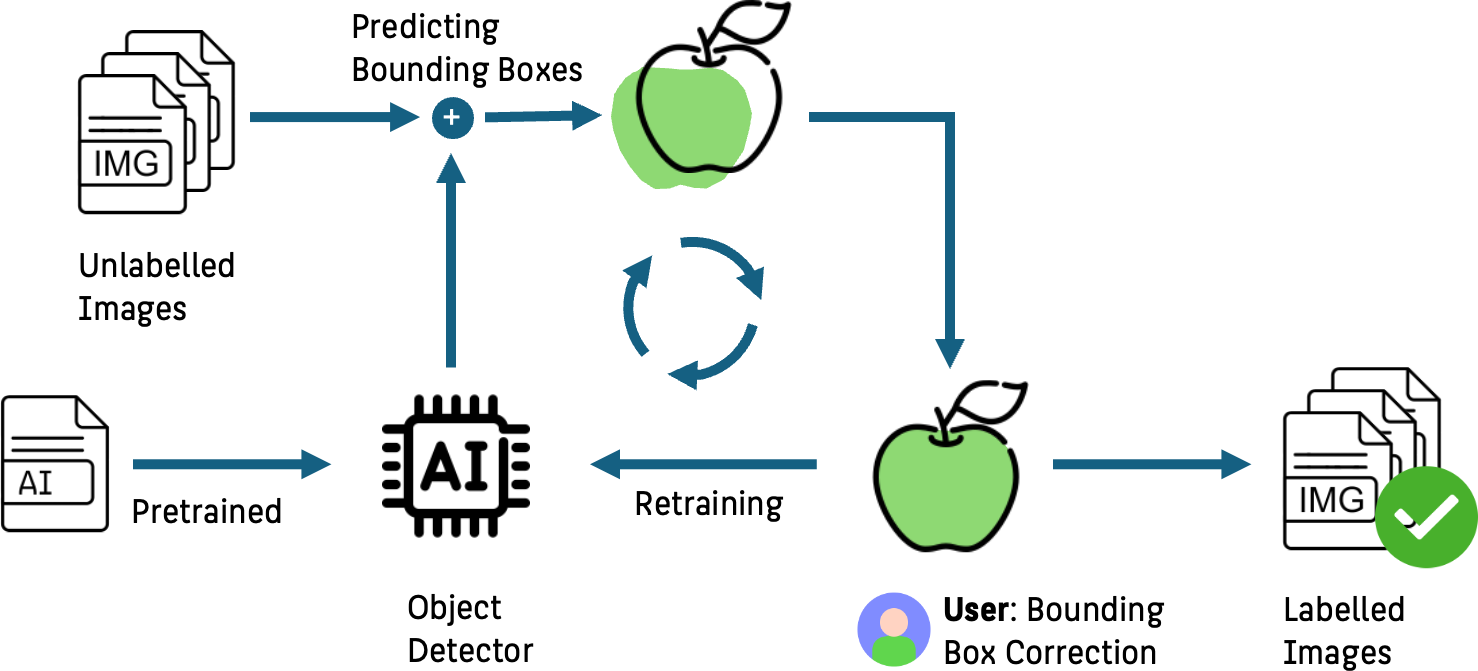}
    \caption{Illustration of the iterative annotation workflow in our platform. The process begins with a pre-trained object detection model predicting bounding boxes on unlabeled images. Users then correct these predictions, and the refined annotations are fed back into the model for incremental improvement. This feedback loop progressively enhances model accuracy, reducing manual annotation effort over time.}
    \label{fig:workflow}
\end{figure}

While accurate, manual annotation is not scalable for large datasets. Consequently, there has been growing interest in methods that integrate machine learning techniques to assist with image labeling, particularly in semi-automatic annotation workflows \cite{degregorio2019semiautomaticlabelingdeeplearning,das2020amazon}. Semi-automatic annotation bridges the gap between fully manual and fully automated annotation by combining the strengths of machine learning and human expertise. Fully automated systems often struggle with rare or ambiguous cases, while semi-automatic methods allow human annotators to correct these mistakes, resulting in higher accuracy and efficiency. This approach promises to significantly improve efficiency without compromising the quality of annotations.

In this paper, we introduce a platform designed to enhance the object detection labeling process through an interactive, feedback-driven loop that iteratively refines model-generated annotations. Our platform allows users to create and manage annotation projects, upload image bundles, and apply a pre-trained Single Shot Detector (SSD) \cite{liu2016ssd} to perform initial object detection and labeling. Users can then manually adjust the generated labels, which are used to incrementally fine-tune the model's performance, creating snapshots for future predictions. This iterative process, which we refer to as semi-automatic annotation, aims to reduce annotation time while maintaining high accuracy.  The workflow of the labeling process is illustrated in Fig.~\ref{fig:workflow}.

We evaluate the platform by conducting quantitative experiments to assess the efficiency of semi-automatic annotations versus fully manual annotation workflows. Our results demonstrate a significant time-saving of up to 53\% without compromising annotation quality. 
The findings presented in this paper provides further quantitative insights and best practices with respect to semi-automatic annotation.

\section{Related Work}


Efficient image annotation is essential for training high-quality object detection models. Manual annotation tools such as LabelMe \cite{russell2008labelme} and VGG Image Annotator \cite{dutta2019via} laid the groundwork by providing interfaces for human annotators to label images. While accurate, manual annotation is time-consuming and not scalable for large datasets. 

Iterative refinement or semi-automatic annotation is indeed a common strategy (see \cite{adhikari2021iterative,schmidt2020advanced,aghdam2019active,munro2021human} just for a few examples). However, a extensive quantiative evaluation of the resulting speed-up as well as suitable best practices are missing in the literature.
Semi-automatic annotation systems, such as Amazon SageMaker Ground Truth \cite{das2020amazon}, combine machine learning with human-in-the-loop corrections, speeding up the annotation process without sacrificing quality. Similarly, Papadopoulos et al. \cite{papadopoulos2017extreme} demonstrated how interactive tools could reduce the time required by allowing users to refine pre-labeled bounding boxes generated by object detectors.

The concept of iterative refinement has also been extensively explored from an active learning perspective, where algorithms select suitable yet unlabeled examples to be annotated or reviewed \cite{settles2009active}. Tools like LabelStudio \cite{Label_Studio} apply this feedback loop in object detection, enabling the model to learn from progressively refined annotations. 

Our platform builds on these advancements by combining the strengths of semi-automatic annotation with an iterative refinement mechanism. This approach reduces the manual annotation workload while continuously improving model accuracy. Our approach can be of course combined with active learning techniques, however, we decided to focus on semi-automatic annotation only.

\section{Platform Design and Implementation}

Building on these previous advancements, our work focuses on the integration of semi-automatic annotation with iterative model refinement to address the limitations observed in manual workflows. The platform leverages semi-automatic annotation, where users correct initial model predictions, and these corrections are used to incrementally improve the model. This process is shown in Fig.~\ref{fig:workflow}. It involves:
\begin{enumerate}
    \item Generating predictions using the SSD model.
    \item Allowing users to refine the annotations.
    \item Using the refined annotations to fine-tune the model incrementally.
\end{enumerate}

\subsection{Object Detection Model}
At the core of the platform's object detection capabilities is a standard SSD model\cite{liu2016ssd}. We employ an SSD300 model pre-trained on the COCO dataset \cite{lin2014microsoft}, chosen for its balance between speed and accuracy. SSD predicts bounding boxes and object classes in a single forward pass, making it ideal for real-time annotation tasks.

\subsection{User Interface}

The platform's user interface (UI) is intuitive and user-friendly, facilitating efficient annotation of large image datasets. It consists of several components guiding users through project creation, annotation, and model fine-tuning.

\subsubsection{Project Management Interface}

\begin{figure}[t!]
    \centering
    \includegraphics[width=0.49\linewidth]{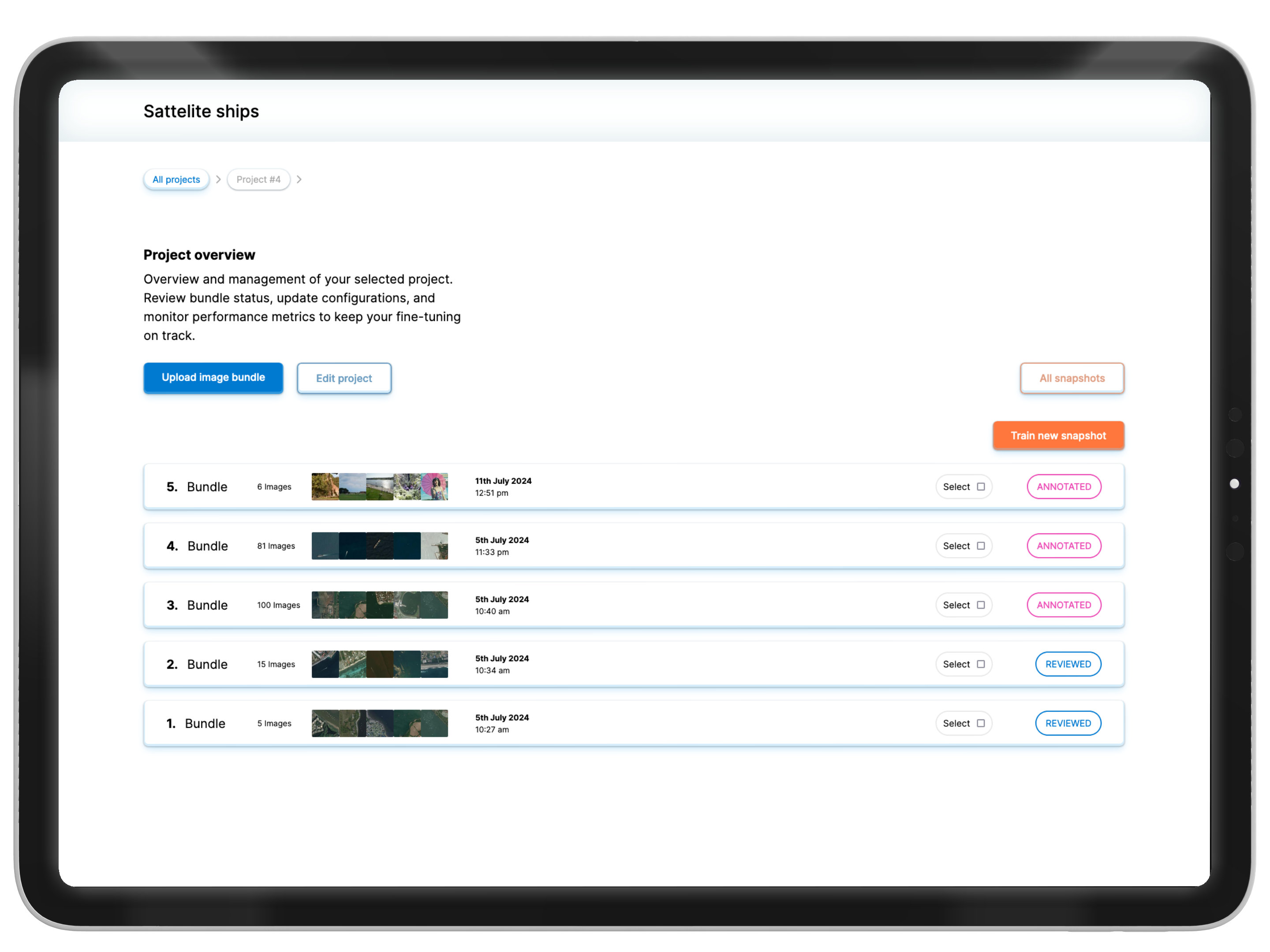}
    \includegraphics[width=0.49\linewidth]{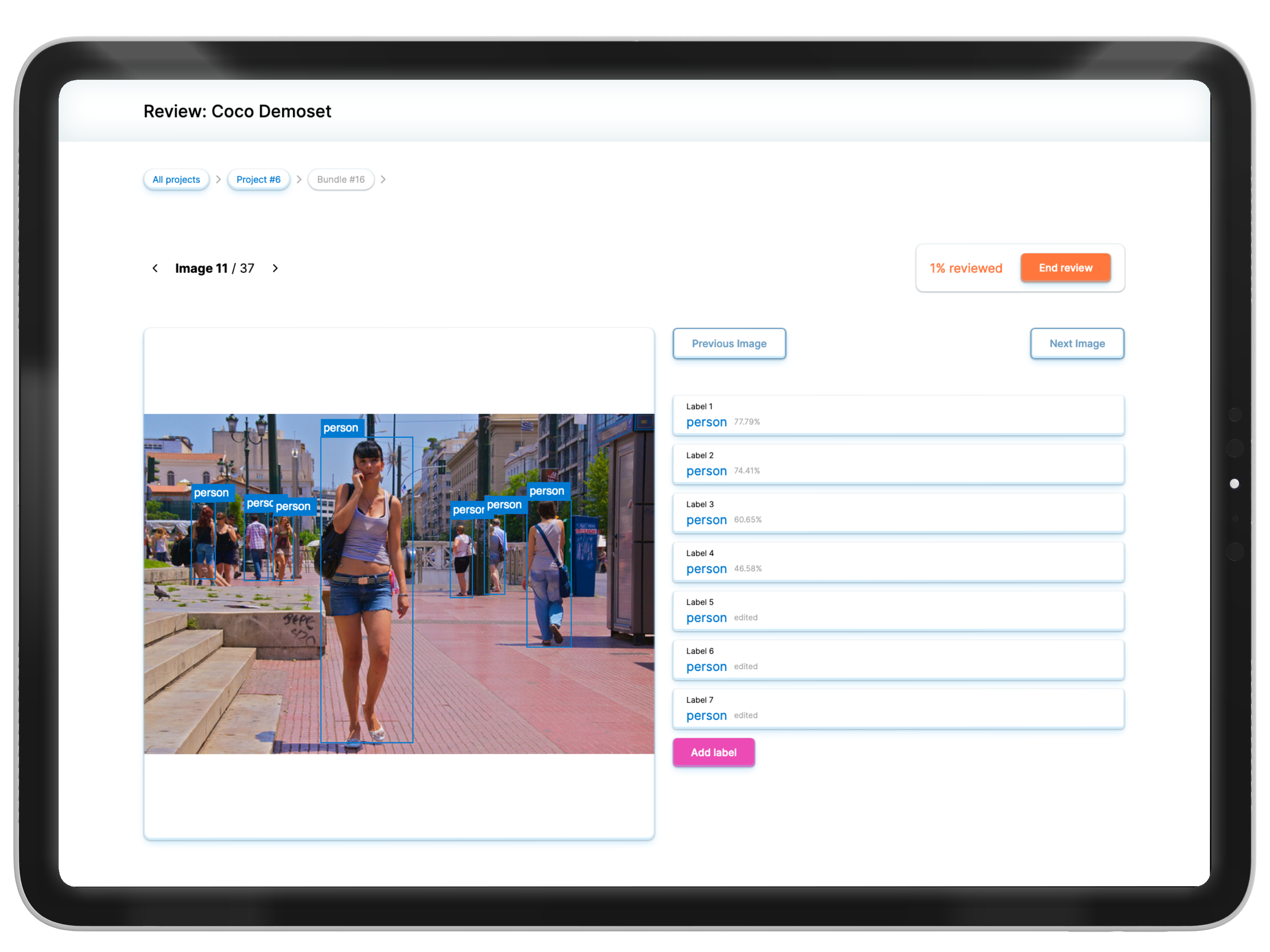}
    \caption{The user interface (UI) of the annotation platform. The left pane shows the project management interface, where users can manage projects, view uploaded image bundles, and monitor the status of model fine-tuning. The right pane displays the annotation editor, where users can review and adjust predicted bounding boxes, creating high-quality annotations through a semi-automatic workflow.}
    \label{fig:ui}
\end{figure}

The project management interface allows users to create, manage, and monitor annotation projects. Fig.~\ref{fig:ui} (left) shows a summary of active projects, with key statistics like uploaded image bundles and fine-tuning status. Users can start new projects, upload images, and review progress through detailed metrics.

\subsubsection{Annotation Editor}

The annotation editor enables users to review and refine bounding boxes generated by the model. Fig.~\ref{fig:ui} (right) shows the interface for adjusting, resizing, or deleting boxes. Users can also add new boxes for missed objects, improving annotation quality through real-time feedback.

\subsubsection{Fine-Tuning Setup}

After refining annotations, users initiate model fine-tuning. The user interface allows selection of image bundles, model snapshots, and training configurations. Performance metrics, such as accuracy and loss values, help users track progress and make decisions on further training.

\subsubsection{Snapshot Management}

Snapshot management ensures version control during fine-tuning. Users can view different versions, compare performance, and roll back when needed, aiding experimentation.

\subsubsection{User Interaction and Shortcuts}

The UI supports keyboard shortcuts for actions like navigating images and adjusting labels, improving efficiency. The platform is also optimized for minimal interaction, ensuring that users can complete the annotation and fine-tuning processes with as few clicks as possible.

\section{Quantitative Evaluation}

\subsection{Datasets}
\label{datasets}
\begin{figure}[t!]
    \centering
        \subfloat[Five plants in a row, with one partially cut off at the edge. The image is taken from a relatively large distance.]{\label{dt-mars-1} \includegraphics[width=0.22\textwidth]{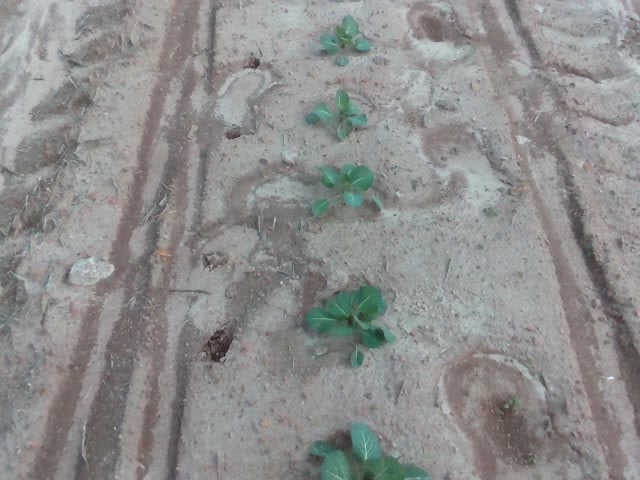}}
        \hfill
        \subfloat[Two plants in a row, with one partially cut off. The camera is positioned very close to the plants.]{\label{dt-mars-2} \includegraphics[width=0.22\textwidth]{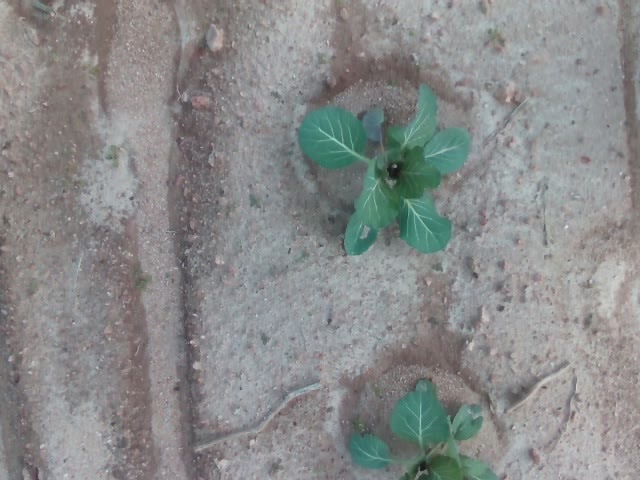}}
        \hfill
        \subfloat[Two plants partially obscured by a shadow with high contrast.]{\label{dt-mars-3} \includegraphics[width=0.22\textwidth]{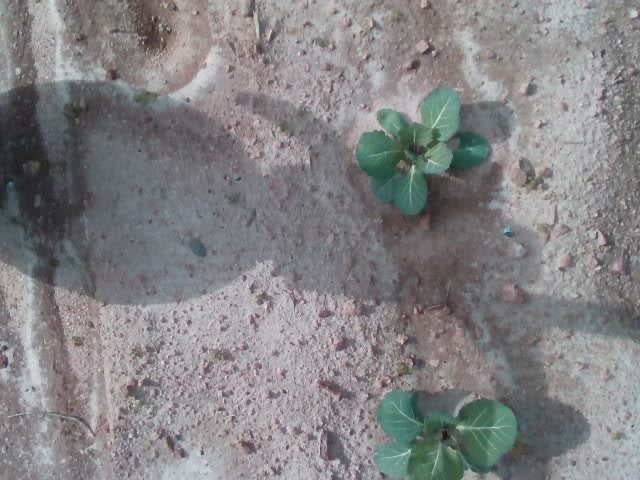}}
        \hfill
        \subfloat[Two plants visible at the edge of the image, with soil disturbances and tracks in the foreground.]{\label{dt-mars-4} \includegraphics[width=0.22\textwidth]{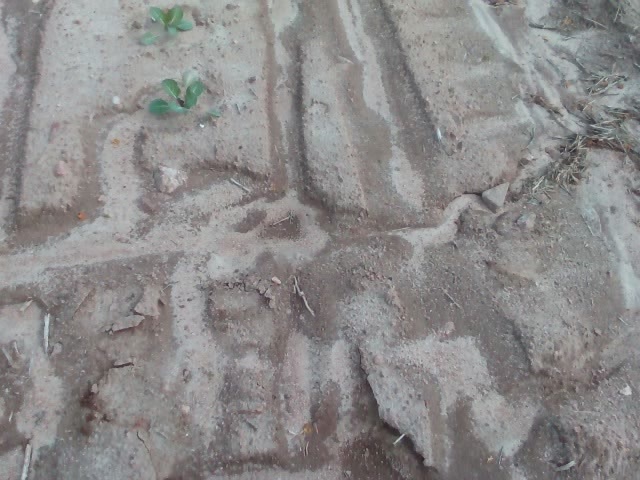}}
    \caption{Example images from the agricultural dataset \cite{dt-mars}, captured by an autonomous agricultural robot. These images illustrate some of the challenging conditions encountered during the annotation process, including varying camera distances, shadows, and partial occlusion of plants.}
    \label{fig:agricultural_dataset}
\end{figure}

The platform was evaluated using the agricultural dataset of \cite{dt-mars}, selected for its challenging conditions that closely simulate real-world agricultural environments. This dataset, consisting of 200 images captured from an autonomous agricultural robot, includes annotated rows of crops from multiple angles and under varying lighting conditions. For our experiments, the dataset was divided into 20 bundles, each containing 10 images, allowing systematic testing of the platform’s adaptability across diverse scenarios. As shown in Fig.~\ref{fig:agricultural_dataset}, these variations—including dense foliage, occlusions, and inconsistent lighting—created a complex environment for the object detection model. Such characteristics enabled us to rigorously assess the platform's performance in handling domain-specific object detection tasks in realistic agricultural monitoring scenarios.

\subsection{Performance Measures}

To evaluate the performance of the platform, we employed several key metrics commonly used in object detection tasks: a) mean intersection over union (IoU), for measuring bounding box overlap between detection and ground-truth. We also calculated the b) F1 score, which is the harmonic mean of precision and recall. Precision and recall were calculated using an IoU threshold of $0.5$ to determine true positives for single detections. The F1 score was computed per image bundle. For the semi-automatic approach, the F1 score was calculated after the automatic labeling was completed, to quantify how much manual intervention was still required to improve accuracy.

Another critical aspect of the platform's evaluation was annotation efficiency. We measured the time required for manual annotation and compared it to the time spent using the semi-automatic annotation process. In addition, we tracked specific user interactions, including the number of bounding boxes created, adjusted, or deleted during the process, allowing for a detailed analysis of the effort saved by using the platform. All measurements were performed on the dataset described in section~\ref{datasets}.

\section{Results}
The experiments conducted in this study demonstrate the effectiveness of the proposed semi-automatic annotation platform. The key findings are summarized as follows:
\begin{enumerate}
    \item The semi-automatic annotation process reduced annotation time by up to 53.82\% compared to manual labeling, especially in later iterations as the model's predictions improved.
    \item User interaction effort, significantly decreased in the semi-automatic process, with nearly no manual box creation needed by the 20th bundle.
    \item F1 scores for semi-automatic annotations matched or exceeded the quality of manual annotations, ensuring that the efficiency gains did not compromise accuracy.
\end{enumerate}

\begin{figure}[t]
    \centering
    \includegraphics[width=1\linewidth]{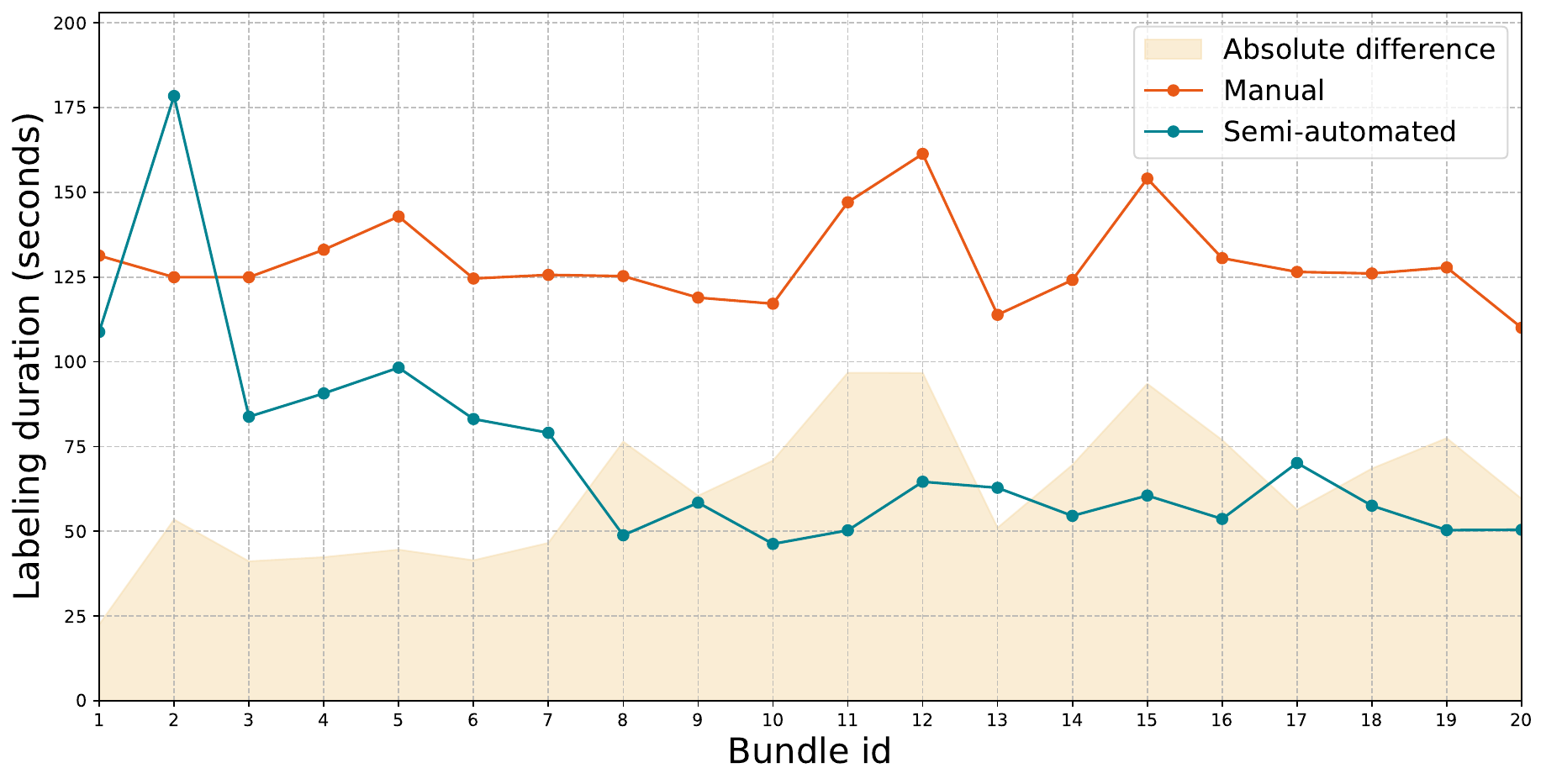}
    \caption{Comparison of the average annotation time per bundle (in seconds) between manual and semi-automatic processes. The results show a significant time reduction using the semi-automatic approach, particularly in later bundles as the model's accuracy improves, reducing the need for manual intervention.}
    \label{fig:time_per_bundle}
\end{figure}

\subsection{Efficiency Gains}
A key objective of this work was to reduce the time required for object annotation. Our experiments, show significant time savings using the semi-automatic annotation platform compared to manual labeling. Fig.~\ref{fig:time_per_bundle} provides a comparison of the annotation duration across different bundles.
It shows, that the semi-automatic process reduces the time spent on annotation by up to 53.82\%, with the most significant reductions observed in later bundles, where the model's accuracy improved. So as the model becomes more refined, user intervention decreases.

\subsection{Interaction Time Reduction}
\begin{figure}[t]
    \centering
    \includegraphics[width=1\linewidth]{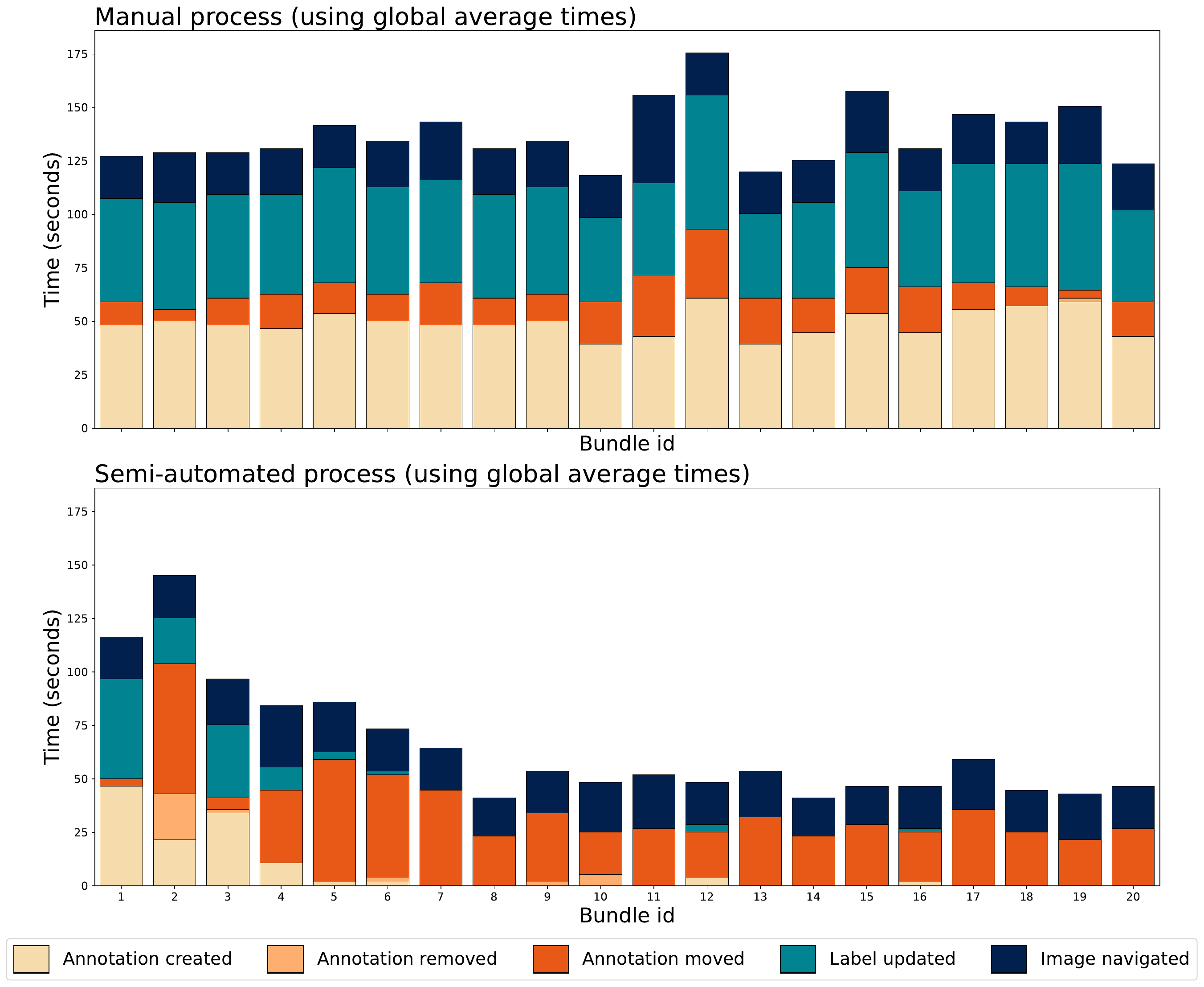}
    \caption{User interaction time metrics comparing the time required for creating, adjusting, or removing bounding boxes in manual and semi-automatic workflows.}
    \label{fig:new_bundle_global_avg_n_10}
\end{figure}

In addition to overall time savings, we measured the time required for individual user interactions, such as creating, adjusting, or removing bounding boxes. Fig.~\ref{fig:new_bundle_global_avg_n_10} illustrates the time taken for each interaction in both manual and semi-automatic workflows. The results show that the semi-automatic process required less time on almost every bundle, with up to 67\% less time needed for the creation of bounding boxes compared to the manual approach.

\subsection{Model Performance}
\begin{figure}[t!]
    \centering
        \includegraphics[width=1\linewidth]{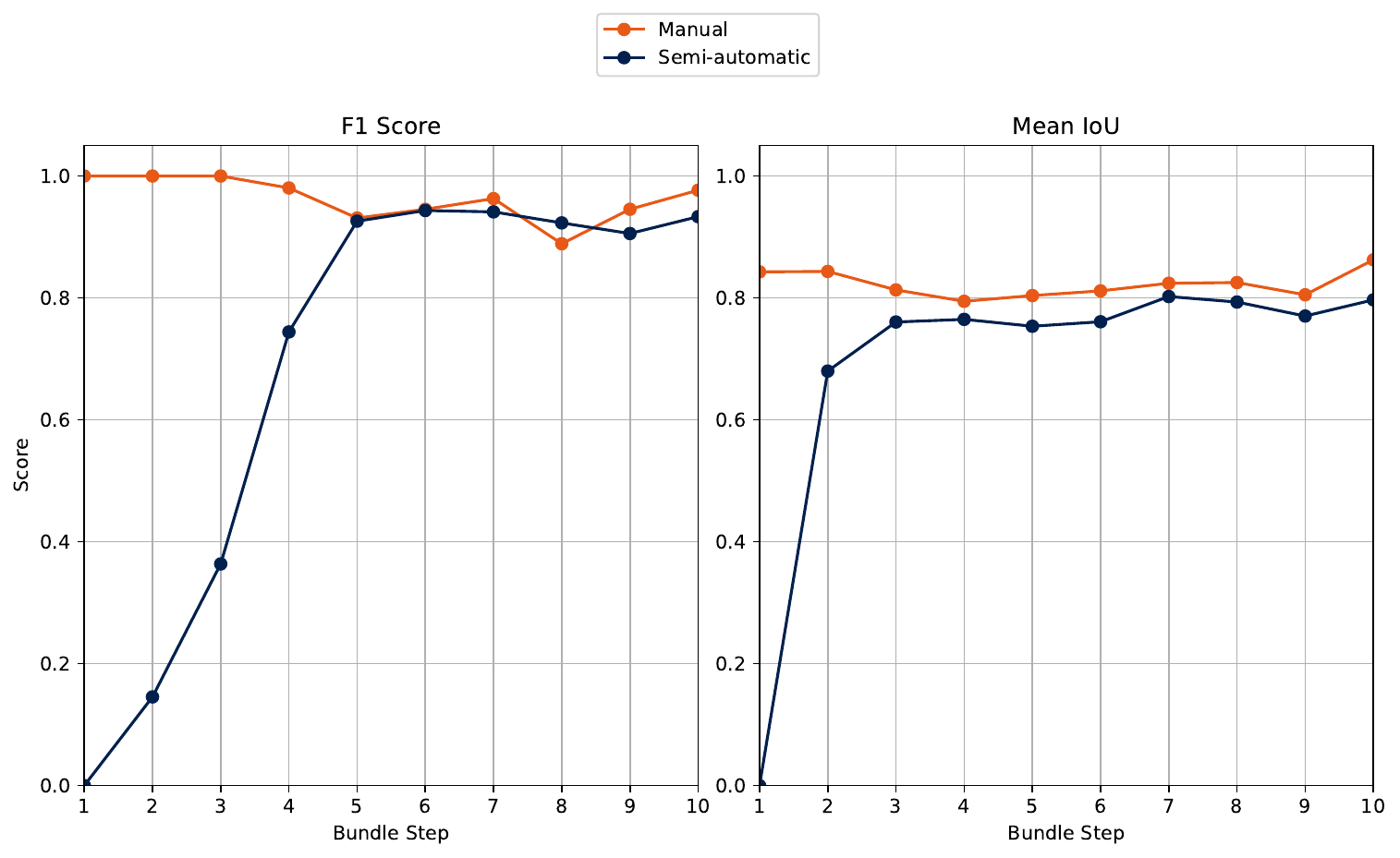}
        \caption{F1 score (left) and mean Intersection over Union (IoU) (right) over multiple iterations of annotation. The analysis is shown up to bundle step 10, because the semi-automatic approach achieved comparable performance to the manual approach by bundle 5, with no further significant divergence in subsequent bundles.}
        \label{fig:performance_curves}
\end{figure}

 As shown in Fig.~\ref{fig:performance_curves} (left), F1 score improved consistently over the course of multiple feedback loops, with the semi-automatic approach achieving comparable performance to the manual approach after just 5 iterations. This indicates that the semi-automatic method significantly reduces the need for manual corrections and can function as a fully automated process after sufficient iterations.

\subsection{Annotation Quality}
In contrast to detection performance, annotation quality was assessed using Intersection over Union (IoU), a widely recognized metric for object detection tasks. IoU measures the overlap between the predicted bounding box and the ground truth bounding box, providing a value between 0 and 1. A higher IoU indicates better alignment between the predicted bounding boxes and the actual objects in the image, thus reflecting higher annotation accuracy.

In this study, the mean IoU was calculated per image bundle, allowing us to track the improvements in annotation quality over the iterative annotation steps. For the semi-automatic approach, mean IoU was computed after the automatic labeling, to assess how accurately the model was able to label objects without requiring further manual adjustments.

The results indicate a steady increase in mean IoU from 0.68 in the first bundle to 0.83 in the final bundle, as illustrated in Fig.~\ref{fig:performance_curves} (right). The comparison shows that automatically generated bounding boxes quickly catch up to the accuracy of manual annotations, with the final iterations producing IoU scores that are on par with fully manual approaches. Eventually, the semi-automatic approach transitions into a fully automated process as the model becomes capable of producing high-quality labels independently.

\section{Conclusion}
In this work, we presented a feedback-driven, semi-automatic annotation platform designed to iteratively improve object detection models. The primary goal was to evaluate the efficiency of such a platform in reducing the time and effort required for creating high-quality labeled datasets compared to traditional manual annotation methods. Through the integration of user feedback and iterative model fine-tuning, we demonstrated that the proposed system achieves significant gains in annotation efficiency without compromising accuracy.

Our experimental results show that the semi-automatic annotation process can reduce annotation time by up to 53.82\% compared to manual labeling, while maintaining or even improving the quality of the annotations. Intersection over Union (IoU) showed consistent improvements across multiple iterations, indicating that the platform effectively enhances the model’s performance over time. These results validate the effectiveness of the semi-automatic approach, especially for large-scale datasets, where manual annotation becomes impractical.
The ability to provide feedback to the model in a dynamic, iterative manner highlights the importance of a human-in-the-loop approach, where machine learning techniques complement human expertise to accelerate the dataset creation process.

\section*{Funding Acknowledgment}

This research was funded by the Deutsche Forschungsgemeinschaft (DFG, German Research Foundation) – Project Number \texttt{528483508}.

%
%
%
\bibliographystyle{splncs04}
\bibliography{paper}

\end{document}